\documentclass[conference]{IEEEtran}

\usepackage{graphicx}
\usepackage{rotating}
\usepackage{subfigure}
\usepackage{multirow}
\usepackage{glossaries}
\usepackage{color}
\usepackage{epstopdf}
\usepackage{listings}
\usepackage{amsmath}
\usepackage{algorithm2e}

\newacronym{6top}{6top}{6tisch Operation Sublayer}
\newacronym{6TiSCH}{6TiSCH}{IPv6 over the TSCH mode of IEEE 802.15.4}

\newacronym{AGV}{AGV}{Automated Guided Vehicle}
\newacronym{AP}{AP}{Allocation Policy} 
\newacronym{AQ}{AQ}{Autonomous Quadcopter}
\newacronym{ASAP}{ASAP}{As-Soon-As-Possible Schedule Another Parent}
\newacronym{ASN}{ASN}{Absolute Slot Number}
\newacronym{ASV}{ASV}{Autonomous Surface Vehicle}
\newacronym{AUV}{AUV}{Autonomous Underwater Vehicle}
\newacronym{ABP}{ABP}{Activation By Personalization}
\newacronym{ACK}{ACK}{Acknowledgment}
\newacronym{ADR}{ADR}{Adaptive Data Rate}
\newacronym{AES}{AES}{Advanced Encryption Standard}
\newacronym{AMQP}{AMQP}{Advanced Message Queuing Protocol}
\newacronym{API}{API}{Application Programming Interface}
\newacronym{AppSKey}{AppSKey}{Application session key}
\newacronym{api}{API}{Application Programming Interfaces}

\newacronym{BW}{BW}{Bandwidth}

\newacronym{CSN}{CSN}{Community Seismic Network}
\newacronym{CRC}{CRC}{Cyclic Redundancy Check}
\newacronym{CH}{CH}{Cluster Head}
\newacronym{CSS}{CSS}{Chirp Spread Spectrum}
\newacronym{CCTV}{CCTV}{Closed Circuit TeleVision}
\newacronym{CEA}{CEA}{Cell Estimation Algorithm}
\newacronym{CIR}{CIR}{Channel Impulse Response}
\newacronym{CoAP}{CoAP}{Constrained Application Protocol}
\newacronym{CoRE}{CoRE}{Constrained RESTful Environments}
\newacronym{CR}{CR}{Cognitive Radio}
\newacronym{CSMA/CA}{CSMA/CA}{Carrier Sense Multiple Access with Collision Avoidance}
\newacronym{CTD}{CTD}{Conductivity, Temperature, Depth}

\newacronym{D2D}{D2D}{Device-to-Device}
\newacronym{DAG}{DAG}{Directed Acyclic Graph}
\newacronym{DAO}{DAO}{Destination Advertisement Object}
\newacronym{DETB}{DETB}{Dynamic Energy \& Traffic Balance}
\newacronym{DGAC}{DGAC}{Direction Générale de l'Aviation Civile}
\newacronym{DIO}{DIO}{Destination Information Object}
\newacronym{DNS}{DNS}{Domain Name System}
\newacronym{DODAG}{DODAG}{Destination Oriented DAG}
\newacronym{DSSS}{DSSS}{Direct Sequence Spread Spectrum}
\newacronym{DevAddr}{DevAddr}{End-device address}
\newacronym{DFT}{DFT}{Discrete Fourier Transform}
\newacronym{dry}{DRY}{Don't Repeat Yourself}

\newacronym{ePFC}{ePFC}{extended Potential Field Controller}
\newacronym{EAM}{EAM}{Earthquake Alert Message}
\newacronym{EDP}{EDP}{Environmental Data Packet}
\newacronym{E2E}{E2E}{End-to-End}
\newacronym{EASA}{EASA}{European Aviation Safety Agency}
\newacronym{ENAC}{ENAC}{Ente Nazionale per l'Aviazione Civile}
\newacronym{EOF}{EOF}{End Of Frame}

\newacronym{FAA}{FAA}{Federal Aviation Administration}
\newacronym{FANET}{FANET}{Flying Ad-hoc NETwork}
\newacronym{FFD}{FFD}{Fully Functioned Device}
\newacronym{FSO}{FSO}{Free Space Optical}
\newacronym{FEC}{FEC}{Forward Error Correction}
\newacronym{FNwkSIntKey}{FNwkSIntKey}{Forwarding network session integrity key}
\newacronym{FPU}{FPU}{Floating Point Unit}

\newacronym{GA}{GA}{Genetic Algorithm}
\newacronym{GWMP}{GWMP}{Gateway Message Protocol}
\newacronym{GIS}{GIS}{Geographic Information System}
\newacronym{GPS}{GPS}{Global Position System}
\newacronym{gui}{GUI}{Graphic User Interface}

\newacronym{HTTP}{HTTP}{HyperText Transfer Protocol}
\newacronym{KPI}{KPI}{Key Performance Index}
\newacronym{KPIs}{KPIs}{Key Performance Indices}

\newacronym{ICN}{ICN}{Information-Centric Networking}
\newacronym{ICT}{ICT}{Information and Communication Technologies}
\newacronym{IETF}{IETF}{Internet Engineering Task Force}
\newacronym{IIoT}{IIoT}{Industrial Internet of Things}
\newacronym{IMC}{IMC}{Internal Model Control}
\newacronym{IMU}{IMU}{Inertial Measurement Unit}
\newacronym{IoD}{IoD}{Internet of Drones}
\newacronym{IoT}{IoT}{Internet of Things}
\newacronym{IP}{IP}{Internet Protocol}
\newacronym{IPv6}{IPv6}{Internet Protocol version 6}
\newacronym{ISM}{ISM}{Industrial, Scientific and Medical}
\newacronym{ITS}{ITS}{Intelligent Transportation System}
\newacronym{I2C}{I2C}{Inter-Integrated Circuit}
\newacronym{IDE}{IDE}{Integrated Development Environment}
\newacronym{IPv4}{IPv4}{Internet Protocol version 4}

\newacronym{JSON}{JSON}{JavaScript Object Notation}

\newacronym{LoRa}{LoRa}{Long Range}
\newacronym{LPWA}{LPWA}{Low Power Wide Area}
\newacronym{LSTM}{LSTM}{Long-Short Term Memory}
\newacronym{LLN}{LLN}{Low-power Lossy Network}
\newacronym{LLSF}{LLSF}{Low-power Lossy Network}
\newacronym{LoRaWAN}{LoRaWAN}{Long Range Wide Area Network}
\newacronym{LoS}{LoS}{Line of Sight}
\newacronym{LPWAN}{LPWAN}{Low Power Wide Area Network}
\newacronym{LTE}{LTE}{Long-Term Evolution}
\newacronym{lstm}{LSTM}{Long Short-Term Memory}

\newacronym{M2M}{M2M}{Machine-to-Machine}
\newacronym{MAC}{MAC}{Medium Access Control}
\newacronym{MANET}{MANET}{Mobile Ad-hoc NETwork}
\newacronym{MCU}{MCU}{Micro-Controller Unit}
\newacronym{MSPRT}{MSPRT}{Multi Sequential Probability Ratio Test}
\newacronym{MTC}{MTC}{Machine-Type Communication}
\newacronym{mcMTC}{MTC}{mission critical Machine-Type Communication}
\newacronym{MTU}{MTU}{Maximum Transmission Unit}
\newacronym{MEMS}{MEMS}{Micro Electro-Mechanical Systems}
\newacronym{MIC}{MIC}{Message Integrity Code}
\newacronym{MMI}{MMI}{Modified Mercalli Intensity}
\newacronym{MPwise}{MPwise}{Multi-Parameter Wireless Sensing System}
\newacronym{MQTT}{MQTT}{Message Queue Telemetry Transport}

\newacronym{NB-IoT}{NB-IoT}{NarrowBand IoT}
\newacronym{NFC}{NFC}{Near Field Communication}
\newacronym{NwkKey}{NwkKey}{Network session key}
\newacronym{NwkSEncKey}{NwkSEncKey}{Network session encryption key}
\newacronym{NLoS}{NLoS}{Non-Line of Sight}

\newacronym{O-QPSK}{O-QPSK}{Offset-Quadrature Phase-Shift Keying}
\newacronym{OS}{OS}{Operating System}
\newacronym{OTF}{OTF}{On The Fly}
\newacronym{OTAA}{OTAA}{Over-The-Air Activation}
\newacronym{oop}{OOD}{Object Oriented Design}

\newacronym{PGA}{PGA}{Peak Ground Acceleration}
\newacronym{PGD}{PGD}{Peak Ground Displacement}
\newacronym{PGHA}{PGHA}{Peak Ground Horizontal Acceleration}
\newacronym{PGV}{PGV}{Peak Ground Velocity}
\newacronym{PGVA}{PGVA}{Peak Ground Vertical Acceleration}
\newacronym{PLANE}{PLANE}{Python Library for simulating unManNed vehiclEs}
\newacronym{pub/sub}{pub/sub}{publish/subscribe}
\newacronym{P2P}{P2P}{Peer-to-Peer}
\newacronym{PCC}{PCC}{Path Computation Client}
\newacronym{PCE}{PCE}{Path Computation Element}
\newacronym{PCEP}{PCEP}{Path Computation Element Protocol}
\newacronym{PDR}{PDR}{Packet Delivery Ratio}
\newacronym{PHY}{PHY}{Physical Layer}
\newacronym{PID}{PID}{Proportional-Integral-Derivative}
\newacronym{PLA}{PLA}{PolyLactic Acid}
\newacronym{PLR}{PLR}{Packet Loss Ratio}
\newacronym{PxIMU}{PxIMU}{Pixhawk Inertial Measurement Unit}

\newacronym{RFD}{RFD}{Reduced Functioned Device}
\newacronym{RMM}{RMM}{Random Mobility Model}
\newacronym{ROLL}{ROLL}{Routing Over Low power and Lossy networks}
\newacronym{ROS}{ROS}{Robot Operating System}
\newacronym{ROV}{ROV}{Remotely Operated underwater Vehicle}
\newacronym{RPL}{RPL}{Routing Protocol for Low-power and Lossy networks}
\newacronym{RTT}{RTT}{Round Trip Time}
\newacronym{RSSI}{RSSI}{Received Signal Strength Indication}
\newacronym{RAN}{RAN}{Rete Accelerometrica Nazionale}
\newacronym{RNN}{RNN}{Recurrent Neural Networks}

\newacronym{QoE}{QoE}{Quality of Experience}
\newacronym{QoL}{QoL}{Quality of Link}
\newacronym{QoS}{QoS}{Quality of Service}

\newacronym{rnn}{RNN}{Recurrent Neural Network}

\newacronym{SF}{SF}{Spreading Factor}
\newacronym{SNwkSIntKey}{SNwkSIntKey}{Serving network session integrity key}
\newacronym{SAIN}{SAIN}{Space-Air Integrated Network}
\newacronym{SDN}{SDN}{Software Defined Network}
\newacronym{SFD}{SFD}{Start of Frame Delimiter}
\newacronym{SF0}{SF0}{Scheduling Function Zero}
\newacronym{SINR}{SINR}{Signal to Interference-plus-Noise Ratio}
\newacronym{SLAM}{SLAM}{Simultaneous Localization And Mapping}
\newacronym{SNR}{SNR}{Signal-to-Noise Ratio}
\newacronym{SPI}{SPI}{Serial Peripheral Interface}

\newacronym{TCP}{TCP}{Transmission Control Protocol}
\newacronym{TDMA}{TDMA}{Time Division Multiple Access}
\newacronym{TI}{TI}{Texas Instruments}
\newacronym{TSCH}{TSCH}{Time Slotted Channel Hopping}

\newacronym{UDP}{UDP}{User Datagram Protocol}
\newacronym{UASN}{UASN}{Underwater Acoustic Sensor Network}
\newacronym{UAV}{UAV}{Unmanned Aerial Vehicle}
\newacronym{UGV}{UGV}{Unmanned Ground Vehicle}
\newacronym{UWSN}{UWSN}{Underwater Wireless Sensor Network}
\newacronym{USV}{USV}{Unmanned Surface Vehicle}
\newacronym{UUV}{UUV}{Unmanned Underwater Vehicle}
\newacronym{UTM}{UTM}{Unmanned Aerial System Traffic Management}
\newacronym{UNB}{UNB}{Ultra Narrow Band}

\newacronym{V2I}{V2I}{Vehicle-to-Infrastructure}
\newacronym{V2V}{V2V}{Vehicle-to-Vehicle}
\newacronym{VANET}{VANET}{Vehicular Ad-hoc NETwork}
\newacronym{VLC}{VLC}{Visible Light Communication}

\newacronym{WG}{WG}{Working Group}
\newacronym{WSN}{WSN}{Wireless Sensor Network}

\newacronym{XSF}{XSF}{X4 Stationnary Flyer}

\begin{document}

\title{PLANE: An Extensible Open Source Framework for modeling the Internet of Drones}

\author{\IEEEauthorblockN{Pietro Boccadoro\IEEEauthorrefmark{1}, Angelo Cardellicchio}
\IEEEauthorblockA{\IEEEauthorrefmark{1}Dep. of Electrical and Information Engineering (DEI), Politecnico di Bari, Bari, Italy,\\ Email:name.surname@poliba.it}}
\maketitle

\begin{abstract}
\textit{\gls{PLANE}} is an open source software module, written in Python, that focuses on \glspl{UAV}, on their movements and on the mechanics of flight, thus devoting particular attention to the equations that describe drones' movement.
In the context of the \gls{IoD}, the module can be widely used for the study of the mutual control of position/coordination in scenarios in which drones may find obstacles, as it happens in densely populated urban scenarios.
Emphasis is put on ease of use, performance evaluation, documentation, and \gls{API} consistency. The software tool has minimal dependencies and is distributed under MIT License. Source code, binaries, and documentation can be downloaded from GitHub.
\end{abstract}


\IEEEpeerreviewmaketitle

\section{Introduction}\label{sec:intro}
The \glspl{UAV}, also known as \textit{drones}, are flying objects with autonomous driving capabilities that are able to carry out a number of different tasks, from simple flight operations (i.e. landing and take-off) to more complex ones \cite{yang2017multi}. The complexity rises up to those operations that are needed in particular operative scenarios;
their use, in fact, spreads from precision farming to emergency management in places where human intervention is difficult, if not impossible \cite{shakhatreh2018unmanned}.
As a further example, drones are used in package transportation/delivery, in photogrammetry and in the study of cement industry and civil engineering,
in the monitoring of coastal areas, analysis and quality control of wooded areas, fire prevention and complete environmental measurement, thanks to several integrated sensors (e.g. temperature and humidity), as well as infrared and video cameras \cite{rice2016drone,Nex2014}.

One of the interesting features drones have is related  the possibility to be remotely controlled, thanks to dedicated interface that allow a pilot who has never taken off to steer the drone.
However, drones can also move autonomously, adjusting flight parameters such as the speed of their engines, attitude and direction, moving along predefined paths \cite{Sahingoz2013}\cite{Belkadi2017}, generally composed by a series of established points (i.e., way-points) which can also be functionally defined for triggering the execution of particular application-driven tasks.

The set of operations a drone has to complete to fly is very complex and foresees activation of engines, variation of their speed and, consequently, different types of maneuver. In the flight mechanics, a drone, which is part of the family of vertical take-off aircrafts, is regulated by a number of variables, such as throttle, inclination, roll and pitch, which can be the result of the controlled variation of the speed of rotation of each motor.
The dynamics of the flight of a drone responds to equations that have already been discussed in the literature \cite{yang2017multi}.
It is worth specifying that drones vary in size and weight. These differences lead to variable maximum transportable weights, rather than different flight autonomy. It may also lead to different maximum reachable altitudes, the resistance to the wind and to atmospheric elements adverse to the flight such as rain, wind, hail and snow.
Even though such knowledge are available, the need arises to create a software that abstracts the aforementioned parameters while embracing the dynamics of flight, thus creating widely descriptive models.

The value proposition of the proposed solution lies in the ability to simulate all the operations that a drone must necessarily carry out to fly.
The realized module is already able to handle coordinates and positions together with engines operations.
The \gls{PLANE} simulator is realized in word scripting language, which allows it to be simple and effective, modular and extensible\footnote{The software tool has minimal dependencies and is distributed under MIT License. Source code, binaries, and documentation can be downloaded from GitHub - https://github.com/anhelus/plane.}. These features leverages the creation of objects identified by classes to which it is possible to associate properties and technical and functional specifications.
Moreover, even the closest solution herein cited is not proposed as an open source solution, left free for further modifications.

The remainder of the present contribution is organized as follows: Section \ref{sec:related} describes the related works. Section \ref{sec:design} proposes the design criteria together with the involved technologies.
Section \ref{sec:future} discusses the future development directions. Finally, Section \ref{sec:conclusions} concludes the work and draws future possibilities.

\section{Related Works}\label{sec:related}
Drones are robotic systems able to fly and their movement can be considered as the result of the complex system of rotors that allows take-off, vertical flight and landing. The dynamics of the flight of a drone can be represented by mathematical equations that are related to physics laws, specifically dynamics and mechanics \cite{yang2017multi}.
When turning from theory to practice, several problems may occur in terms of both mechanics and stability of the system once on a flight mission. To properly understand flight-related problems, it is of relevance to simulate both the mechanics and the stability of the system once on a flight mission.
To this aim, software solutions may be used to create demos and mock-ups to check if the results of the model under analysis still remains stable and/or reliable over long periods of time. Simulations may also account for relevant results in extensive experimental campaigns on physical systems.
The rationale for the employment of software aids is also motivated by the fact that the prototype of a real system may suddenly meet limits in terms of flight capacity and autonomy.
As a matter of fact, such problems may occur as the optimization is still ongoing and the battery supply discharge may be lowered.
The modeling of \glspl{UAV} could be strongly enhanced by taking deep learning into consideration. The rationale for this assumption is related to the fact that it has shown a great versatility and noticeable results in solving a variety of problems connected to robots operations, such as rovering and navigation, path planning, localization, and control \cite{carrio2017review}.
Deep learning is extremely useful when features have to be extracted by complex data-sets resulting from the acquisition from real environments, which is part of the core business of \glspl{UAV} operations, from security/surveillance to warehouse management.
\cite{carrio2017review} also discusses the challenges in one of the major concerns about \gls{UAV}, i.e., autonomy. In particular, \gls{UAV} platforms still have constraints in flight endurance. These are generally solved by shrinking sizes and/or weights and/or power consumption and payloads. Moreover, on-board processing capabilities are generally limited too.
These challenges suggest that the development of advances embedded hardware technologies could lead the way toward efficient deep learning-based architectures.

In general, the contributions proposed so far are three-folded:
\begin{itemize}
	\item \textbf{Remote Control Systems} \cite{FDH17} refers to a class of software solutions that act as an interface between the operator and the automated vehicle. This software emulates the presence of the pilot on-board of the \gls{UAV} to guarantee that every flight operation corresponds to something that the pilot meant.
	
	\item \textbf{Autonomous Control Systems} \cite{CMB+16}, related to a different class of flight problems that aims at supporting autonomous flying objects. These software solutions implement both control and driving logic.
	
	\item \textbf{Flight Simulators} \cite{AMB+12} a possibility that refers to the simulation of the flight for training pilots. Indeed, it is not specifically thought for drones and does not specifically relates to their autonomy and/or self-driving capabilities. 
\end{itemize}

In \cite{6731982} it is proposed the implementation of a 4-rotor control system\footnote{Note that it does not fit all real drones, as a number of them are based on six-rotors.} solution, verified using MATLAB Simulink. It is worth remarking that these studies are specifically designed for 4-rotors systems.

The work presented in \cite{8207649} addresses moving targets, with a specific reference to the knowledge of their mobility patterns. This work is based on a pragmatical approach that facilitates the implementation of the proposals in the context of real-time coverage problems it mainly aims at supporting applications based on mobile target tracking. Unfortunately, the work does not propose results for long flight times and/or distances.

\cite{Zemalache2009} describes an automatically on-line Self-Tunable Fuzzy Inference System (STFIS) of a mini-drone, i.e., \gls{XSF}.
The proposal is based on a fuzzy controller based on an on-line optimization routine. In particular, the descent gradient algorithm adapts to unknown situations whereas the empirical knowledge modeling based on fuzzy logic are combined in order to control the engines.
Among the most interesting features that the proposal enables, obstacles avoidance and multi-drone flying are envisaged.

In \cite{Sahingoz2013}, the authors argues that when the number of waypoints arises, together with the number of flying objects in a given area, optimization algorithm are mandatory to lower the computational overhead and the number of useless movements and/or operations carried out by each drone.
The rationale for the employment of multiple \glspl{UAV} is related to the fact that when the mission has multiple objectives, tasks differentiation among the drones in the swarm leads to increasing the efficiency of the system, while lowering the overall operational time.
Therefore, it is designed a path planning algorithm for multi-\gls{UAV} systems. The solution is a \gls{GA}-like and foresees a number of \glspl{UAV} flying at constant altitude in a known environment. In the proposed solution, once the \gls{GA} routine has found a feasible solution, it is smoothed using Bezier curves.
This visionary proposal still has some margins of enhancements since it does not address kinetic problems such as minimum curvature and/or torsion and maximum climb angle.

\cite{PM16} represents a very interesting model-based test for the well-known Ardupilot platform\footnote{http://ardupilot.org/}. Still, the contribution is not a simulator, as it builds a model that addresses core functionalities of autonomous flight in fail-safe conditions. The proposed control loop improves the existing solution leveraging efficient operations and solving some bugs. The proposal is of relevance since it contributes to the development of the platform.

In \cite{GARCIA2018875}, an interesting contribution addresses the functional and behavioral description of the peculiar drones' mechanical components. Nevertheless, the proposal does not specifically solve the same problem. In fact, in this work the drone is not considered as a whole.

All the above suggest that there is the need for a complete framework which can be used for modeling, simulation and experimental evaluations.
The present contribution proposes an incremental point of view on the theme of simulations that takes into account and implements the mathematical framework of such unmanned vehicles.
The \gls{PLANE} simulation tool will be proven as a modular system, expandable and further adaptable solution.

\section{Code Design}\label{sec:design}
\subsection{Requirements}

The first step envisages the study of the basic mechanics upon which a proper representation of the behavior of a drone (as long as a swarm of) may be defined. This representation can be basically traced back to a pipeline of \textit{modeling} and \textit{optimization} techniques. First, a modeling approach is needed to describe the inner behavior of the swarm, along with the relationships between its members. The resulting outcome, which is in the form of a set of highly nonlinear models, should be used as the input of the optimization approach. It allows to define the optimal routes followed by each drone according to a set of mandatory constraints. 

\subsection{Selected tools}

According to the previously defined requirements, and following the \gls{dry} paradigm, \textit{\gls{PLANE}} bases its operation on a set of widely used, stable and open source libraries for data analysis, modeling and optimization.

First, the \textit{SciPy ecosystem} \cite{scipy} can be used as a common ground upon which other tools are built. Specifically, \textit{NumPy} \cite{numpy} offers several functionalities of interest for numeric and matrix computation, while \textit{SciPy} itself offers scientific processing capabilities (e.g. signal processing, statistical and probability analysis, etc.). Furthermore, \textit{Pandas} \cite{pandas} is used as a tool for handling data sources and preprocessing tasks.

As for modeling, two different libraries are used:
\begin{itemize}
	\item \textit{Scikit Learn} \cite{scikit}, employed for both unsupervised and supervised modeling. Specifically, Scikit Learn offers an out-of-the-box implementation of a wide range of machine learning algorithms, which can be used to model the behavior of either single drones or swarms. Furthermore, it can be seamlessly integrated with the other libraries used by \textit{\gls{PLANE}}, and its algorithm benchmarking tools can be used to assess the performance in terms of modeling and optimization.
	
	\item \textit{TensorFlow} \cite{tensorflow} is used for deep learning. Specifically, it is extremely useful as it enables time series modeling and forecasting by means of \gls{rnn} with \gls{lstm} gates \cite{carrio2017review}, which can be useful in modeling the behavior, over time, of both single drones and swarms.
\end{itemize}

Finally, the optimization task is carried out using \textit{OR-Tools} \cite{ortools}, a framework, developed by Google, and specifically aimed at solving several types of optimization problems, such as linear and constraint programming, scheduling and routing. Specifically, \textit{\gls{PLANE}} uses optimization procedures related to drone routing and path finding to find the best route according to several constraints.

\subsection{Design principles}\label{subsec:design_principles}

\textit{\gls{PLANE}} has been designed according to the three main principles: (i) \textit{modularity}, (ii) \textit{extensibility}, and (iii) \textit{developer-oriented} design.

As for the first, \textit{\gls{PLANE}} is composed by several components (i.e., Python modules), which are designed to be self-contained and loosely coupled. Data sharing between modules is guaranteed by specific interfaces. Modularity is directly related to two principles on which \gls{oop} is based, that is, \textit{composition} and \textit{separation of concerns}. Specifically, composition allows to 'build' complex objects (e.g. a drone) by means of simpler ones (e.g. rotors, body and propellers), while separation of concerns establish that each object should be solely responsible for its operation and behavior.

The second principle is \textit{extensibility}. Each module within \textit{\gls{PLANE}} follows the basic pattern defined by the \textit{inheritance} paradigm. This allows to define a base implementation over each type of component (such as the \textit{motion} components) which will be specialized and extended for the specific use-case.

The third design principle is \textit{developer-oriented} design. Specifically, developers should be able to use a commonly defined \gls{api} to access to properties of different objects, therefore simplifying the adoption of the framework.

\subsection{Workflow and structure}
The core \gls{PLANE} workflow for simulating routes followed by drones is described by Algorithm \ref{alg:PLANE_workflow}.

\begin{algorithm}[h]
	\SetAlgoLined
	\KwResult{Simulate the route of an entire swarm}
	initialize swarm\;
	initialize scenario\;
	\For{drone in swarm}{
		compute drone model\;
		compute interactions with other drones\;
		simulate route\;
	};
	\caption{The core workflow followed by PLANE for simulating routes of an entire swarm.}
	\label{alg:PLANE_workflow}
\end{algorithm}

\begin{figure*}[ht]
	\centering
	\includegraphics[width=1\linewidth]{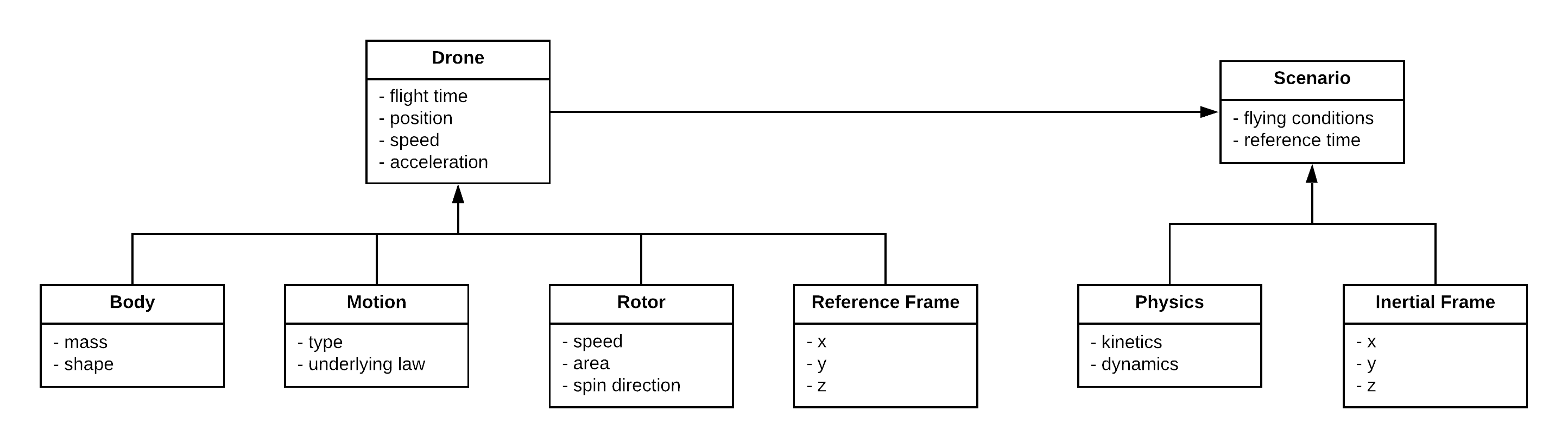}
	\caption{The class diagram of PLANE. It is important to underline that the Drone and the Scenario classes both contribute to the simulation of the overall path of a swarm.}
	\label{fig:PLANE_class_diagram}
\end{figure*}

The workflow starts with the initialization of each drone. This implies that all the values for the required properties of each drone should be defined and set in this step. A drone is modeled by several attributes, related to both its physical properties and its contextual scenario. Specifically, as described in figure \ref{fig:PLANE_class_diagram}, each drone is characterized by the following.

\begin{itemize}
	\item \textit{Flight time, position, speed and acceleration}: these can be defined as \textit{inner} properties for each drone, and are used to describe both the current position and the motion behavior of the drone.
	\item \textit{Body}: this property is modeled by a specific class, which physically define the drone. These information directly influence the inner properties, along with the speed required by each rotor and the position of the drone within the reference frame.
	\item \textit{Motion}: this class models the motion to which the drone is currently subjected in mathematical terms.
	\item \textit{Rotor}: this class models one of the rotors which are responsible for the motion of the drone, through its speed, area and spin direction. Specifically, speed and area define the total thrust along one or more axis, while the spin direction is used to define whether the rotor is spinning in \textit{clockwise} or \textit{counter-clockwise} direction, hence \textit{increasing} or \textit{decreasing} drone's altitude. Each drone may have two or more rotors.
	\item \textit{Reference Frame}: this class represents the inner reference frame of the drone, therefore determining its orientation along the \textit{x}, \textit{y} and \textit{z} axes.
\end{itemize}

An interesting consideration is that PLANE is designed to make each drone belonging to a \textit{set} (that is, a swarm); therefore, a single drone is also conveniently modeled as a set composed by a single element. Once the swarm has been initialized, PLANE creates an instance of the \textit{scenario}, which can be also referred to as \textit{world} or \textit{reference scenario}. Such instance is used to define the context within which the swarm travels.

The scenario is described by means of the following properties:

\begin{itemize}
	\item \textit{Flying conditions and reference time}: these attributes can be also referred to as the \textit{inner} properties of the scenario. Flying conditions model all the contextual and environmental properties which can be found within the world, such as weather, buildings, etc. The reference time is also extremely important, as it is used to synchronize each drone within the swarm, and give a common reference frame to the world.
	\item \textit{Physics}: each scenario embeds a set of physics laws, which are currently modeling kinetics and dynamics, and are defined through a set of constants and mathematical equations.
	\item \textit{Inertial Frame}: the inertial frame is a common reference frame, defined for the whole scenario, and where each drone translates and rotates according to its own reference frame.
\end{itemize}

\section{The road ahead}\label{sec:future}
\textit{\gls{PLANE}} is still under development, hence, several feature will be added to the upcoming releases of the framework. Some of them are hereby listed:
\begin{itemize}
	\item A set of common cases will be designed and implemented, offering a common ground for analysis and algorithm benchmarking.
	\item Flight mechanics will be enriched, supporting the integration of data relative to the environmental context and flying conditions (e.g. weather, buildings and trees, etc.).
	\item Extensive support to geo-referenced data will allow to store and return spatial information on routes, which will be exposed through the GeoJSON format to fed external tools such as Google Maps and QGIS.
	\item Support to data exporting and importing will be included, to allow interoperability with external data analysis tools.
	\item Integration and/or interaction with Network Simulator 3 (ns3)\footnote{https://www.nsnam.org/} \cite{stojmenovic2008sws} will be granted to analyze a drone-based communication network architecture.
	\item A user interface will allow non-technical users to easily perform simulations and test cases on both already implemented and custom scenarios.
	\item The set of metrics for analytics and benchmarking will be extended, including useful metrics such as Baesyan Information Criterio, MAPE, Root Mean Square Error, and more.
	\item An extensive documentation, both as tutorials and concepts reference, will be made available.
\end{itemize}

Finally, all future implementations will be developed according to principles already exposed in section \ref{sec:design}, in order to ensure API consistency and a shared implementation.

\gls{PLANE} can be considered a promise for the future and will certainly support research and development efforts in the \gls{IoD} \cite{Gharibi2016} field. The joint adoption of both \gls{PLANE} and a network simulation tool could leverage simulations to a brand new level.

\section{Conclusions}\label{sec:conclusions}
This work presented \textit{\gls{PLANE}}, an open source, extensible framework for the analysis of drones and swarms. 
\textit{\gls{PLANE}} already allows both model and data driven approaches to the analysis of common scenario.
In the future, the software tool will be extended taking into account flight mechanics and more precise models.
Moreover, the tool will be enriched with analytic and benchmarking routines
Furthermore, the integration with network simulation tools will be studied in details in order to provide more reliable and general purpose software simulation tools.
By the way, the most thrilling research perspective comes from the possibility to jointly adopt \textit{\gls{PLANE}} and a network simulator to evaluate communication effectiveness while drones are effectively moving. Such research perspective may pave the way toward a thorough \gls{IoD} evaluation tool.

\bibliographystyle{IEEEtran}
\bibliography{bibligraphy}

\end{document}